\documentclass[runningheads]{llncs}
\usepackage{graphicx}
\usepackage{comment}
\usepackage{amsmath}
\usepackage{amssymb}
\usepackage{multicol}

\usepackage{color, colortbl}
\definecolor{LightCyan}{rgb}{0.88,1,1}

\definecolor{deepmagenta}{rgb}{0.8, 0.0, 0.8}

\usepackage[first=0,last=9]{lcg}

\usepackage{makecell}
\usepackage{array}
\newcolumntype{P}[1]{>{\centering\arraybackslash}p{#1}}
\newcolumntype{M}[1]{>{\centering\arraybackslash}m{#1}}

\usepackage{subcaption}
\usepackage[toc,page]{appendix}
\usepackage{rsfso}
\usepackage{float}

\usepackage{hyperref}
\hypersetup{
    colorlinks=true,
    linkcolor=blue,
    filecolor=magenta,      
    urlcolor=blue,
    citecolor=red,
}
\usepackage{xcolor}

\DeclareMathOperator*{\argmax}{arg\,max}

\begin{document}
\title{Learning-based vs Model-free Adaptive Control of a MAV under Wind Gust\thanks{\textit{This work was supported by ISblue project, Interdisciplinary graduate school for the blue planet (ANR-17-EURE-0015) and co-funded by a grant from the French\hspace{20px}government under the program "Investissements d'Avenir".}}
}
%
%
\author{Thomas Chaffre\inst{1,2} \and
Julien Moras\inst{3}             \and
Adrien Chan-Hon-Tong\inst{3}     \and
Julien Marzat\inst{3}            \and
Karl Sammut\inst{2}              \and
Gilles Le Chenadec\inst{1} \and Benoit Clement\inst{1,2}
}

\authorrunning{T. Chaffre et al.}
%
\institute{Lab-STICC UMR CNRS 6285, ENSTA Bretagne, Brest, France \email{\{thomas.chaffre,gilles.le\_chenadec\}@ensta-bretagne.org}\and
Centre for Maritime Engineering, Flinders University, Adelaide, SA, Australia \email{\{karl.sammut,benoit.clement\}@flinders.edu.au}\and
DTIS, ONERA, Université Paris-Saclay, 91123 Palaiseau, France
\email{\{julien.moras,adrien.chan\_hon\_tong,julien.marzat\}@onera.fr} }
\maketitle 
\begin{abstract}
Navigation problems under unknown varying conditions are among the most important and well-studied problems in the control field. Classic model-based adaptive control methods can be applied only when a convenient model of the plant or environment is provided. Recent model-free adaptive control methods aim at removing this dependency by learning the physical characteristics of the plant and/or process directly from sensor feedback. Although there have been prior attempts at improving these techniques, it remains an open question as to whether it is possible to cope with real-world uncertainties in a control system that is fully based on either paradigm. We propose a conceptually simple learning-based approach composed of a full state feedback controller, tuned robustly by a deep reinforcement learning framework based on the Soft Actor-Critic algorithm.
We compare it, in realistic simulations, to a model-free controller that uses the same deep reinforcement learning framework for the control of a micro aerial vehicle under wind gust. The results indicate the great potential of learning-based adaptive control methods in modern dynamical systems.
\end{abstract}

\renewcommand{\thefootnote}{1}
\section{Introduction}
Aerial vehicles are exposed to a mixture of perturbations that fluctuate vigorously because of the hazardous environment they are evolving in. They operate over a wide range of speeds and altitudes. Their dynamics are nonlinear and can also be of time-varying nature (as they fly, their mass slowly decreases due to fuel consumption and their center of gravity can greatly vary). Micro Aerial Vehicles (MAVs) are challenging in some other ways. They have to constantly compensate for the distribution of forces that act on their small airframes because of wind and turbulence. These forces can significantly increase when weather conditions change and are also influenced by the velocity and heading of the vehicle. Because they are small, rigid, light, and move slowly, MAVs are highly sensitive to wind gusts even if mild. In addition to the latter unpredictable (external) forces, we face the difficulty of accurately modeling their (internal) dynamics under these conditions. With the development of flight controllers for hypersonic aircraft by NASA in the early $1950$'s \cite{NASAX15}, it was found that conventional linear feedback control strategies (e.g. fixed-gain control \cite{JTerrell}, robust control \cite{CDoyle}, sliding mode control \cite{IUtkin} or fuzzy logic control \cite{RBellman}) are too limited to handle such entire regimes. It is also true for the MAVs context where for a variety of reasons it is almost impossible to manually re-tune the control parameters, these include:
\begin{itemize}
    \item The  uncertainty level of the wind speed is high and can vary more quickly than can be compensated by these types of controllers.
    \item The response at some operating points has to be overly conservative in order to satisfy specifications at other operating points.
    \item The controlled process itself varies significantly during operations (the motors' and propellers' efficiency for example).
\end{itemize}

The field of adaptive control broadly addresses these challenges by granting the control law some flexibility to modify its action based on the process variations. Despite the great advancements conducted in the theoretical area, the expansion of autonomous agents to the real-world is still limited. This is mostly due to the high dimensional problems that often appear when working under real conditions where it usually requires many degrees of freedom to describe the robot state. The lack of basic knowledge of the various natural phenomena taking place in the robot's environment, e.g. wind gust for MAVs or sea current for autonomous underwater vehicles (AUVs) makes it even more difficult to actively control it \cite{Anderson2005} with model-based adaptive methods.

For the past few years, model-free adaptive control methods have attracted significant attention. They exploit strong statistical tools that provide control systems the ability to automatically learn and improve from experience without being explicitly told how to. In particular, model-free adaptive methods based on deep reinforcement learning  (DRL) have shown promising achievements. This was possible especially thanks to the use of deep neural networks to extract physical insights through sensory data (empowered by progress in data collection with high fidelity simulations, cheaper storage drives, faster computers, etc). However, as stated in \cite{Snderhauf2018TheLA}, the extension of DRL techniques for robotic tasks raises serious questions. Compared to other application fields, robots have to interact with a dynamic environment where relevant information about the system is not always accessible or tractable over time.

In their up-to-date thorough investigations~\cite{DulacArnold2020AnEI}, G. Dulac-Arnold \textit{et al.} presented real-world reinforcement learning challenges that are still not resolved, these include \textit{Satisfying Environmental Constraints, High-Dimensional Continuous State and Action Spaces or Multi-Objective Reward Functions}. In addition, the notions of stability (in terms of Lyapunov stability) in the DRL framework is more deeply investigated~\cite{Garcia2015ACS} but the lack of formalism is still often highlighted by the control community as a significant concern. The purpose of this study is to explore to what extent model-free control techniques combined with the classical theory of dynamical systems can be, at least, part of the solution to these great challenges. This is the objective of the learning-based control area \cite{Mouhacine16} which aims to combine the advantages of the aforementioned paradigms (model-based and model-free) into one hybrid control scheme.
%
Therefore, we propose a learning-based adaptive control system composed of a state-feedback control law whose parameters (i.e. gains) are tuned by the Soft Actor-Critic DRL algorithm \cite{haarnojaSoft} using the pole placement method. It is fairly compared, in a realistic simulation setting, with a model-free adaptive method trained  end-to-end with the same DRL framework (as initially applied to robot navigation in~\cite{TChaffre}). The control objective treated is a hexacopter performing a waypoint rallying mission under unknown wind fields.

The paper is organized as follows. Section~\ref{relatedworks} gives details on the theoretical bases of the proposed control strategies and related work. In Section~\ref{background}, we present background material on the simulation and control of MAVs. Section~\ref{method} is dedicated to the description of our contribution, with the application of DRL to the design of model-free and learning-based controllers.
A description of the simulated training and evaluation sessions is provided in Section~\ref{results} with the obtained results. Finally, a thorough analysis of the outcomes and some suggestions to extend the work further are presented in Section~\ref{discussion}.

\section{Related work} \label{relatedworks}
An adaptive controller is essentially a controller that can adjust its own behavior in response to changes in the dynamics of the process and the character of the disturbances~\cite{Astrom2nd}. Since ordinary feedback control also seeks to reduce these undesirable effects, one needs to clearly distinguish between the two. In this paper, we will designate as \textit{adaptive}, a control system that provides the desired system performance by changing its parameters and/or structure in order to reduce the uncertainties effect and to improve the knowledge of the desired system. As mentioned by Nikolai M. Filatov~\cite{FilatovDual}, the goal of adaptive systems is to switch off the adaptation as soon as the system uncertainty is reduced sufficiently and to then use the adjusted controller with a fixed structure and fixed parameters.
Although it is more common in the control community to classify methods based on the nature of the model (e.g. linear vs nonlinear, continuous vs discrete), we classify here adaptive controllers based on their dependence on the mathematical model of the controlled system. This algebraic representation of the dynamics of the system can take several forms such as differential equations, state-space representations, or frequency domain representations. We can thus identify three types of adaptive control methods: totally model-based, totally model-free, and learning-based. Since the approach presented here consists in merging together some model-based controllers and model-free algorithms, in this chapter we will present the basis of the aforesaid techniques and recent applications to the problem of MAVs under unknown wind fields. Then, we will see how reinforcement learning can be used to strengthen control laws, which will lead to the introduction of the proposed learning-based adaptive controller.
\begin{figure}[H]
   \begin{center}
      \includegraphics[scale=0.5]{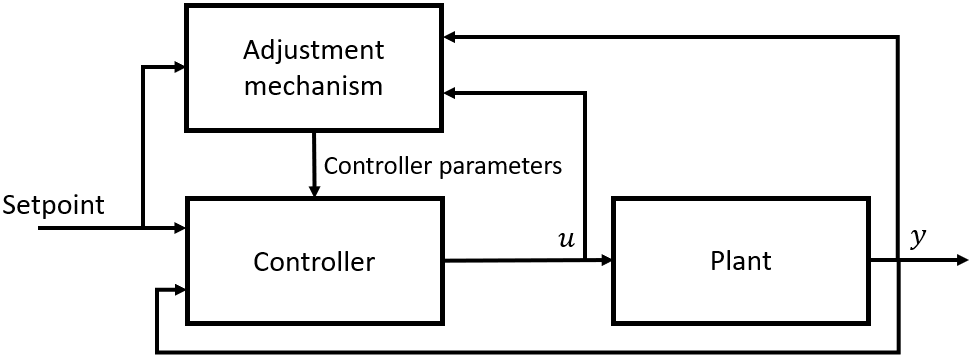}
   \end{center}
   \caption[todo]
   {\footnotesize General block diagram of an adaptive control system. It is composed of two loops: a normal feedback loop with the plant and the controller and a second loop with the parameter adjustment mechanism (which is often slower than the first one).}
   \label{fig:adconntroller}
\end{figure}

\renewcommand{\thefootnote}{2}
\subsubsection{Formulation of the adaptive control problem:\\ \\}
\hspace{-5px}Consider the system described by the following continuous-time state equations:
\begin{equation} \label{adaptproblem}
\begin{split}
&\dot{x}(t)=f(x(t),u(t),p(t), t),\\
&y(t)=h(x(t),u(t), t),
\end{split}
\end{equation}
where $x\in \mathbb{R}^n$ is the system state\footnote{In this paper, $x$ refers to state in terms of state space representation and $s$ in regard to the context of machine learning because although sharing the same title, these entities do not hold the same nature.} ; $u\in\mathbb{R}^{n_u}$ is the control vector ; $p(t)\in\mathbb{R}^p$ is the constant or time-varying vector composed of the unknown or uncertain parameters of the model ; 
$y\in\mathbb{R}^m$ is the output vector of interest. The probability density of the initial values $\mathbb{P}[x(0),p(0)]$ is assumed to be known. We can define the set of outputs and control inputs available at time $t$ as:
\begin{equation}
    \Gamma_t=\{y(t),\dots,y(0),u(t-1),\dots,u(0)\},\hspace{5px}t=0,1,\dots,T-1,\hspace{5px}\Gamma_0=\{y(0)\} 
\end{equation}

\noindent The control performance index can have the following form:
\begin{equation} \label{perfindex}
    J(t)=\frac{1}{t}\int_0^t\Bar{e}^2(\tau)d\tau,\hspace{5px}\Bar{e}(t)=w(t)-y(t)
\end{equation}
where $w(t)$ is the targeted set point. The general problem of adaptive control consists in finding the control policy $u(t)=u_t(\Gamma_t)\in\Bar{\Omega}_t$ that minimizes the performance index \eqref{perfindex} for the system described by  \eqref{adaptproblem} with $\Bar{\Omega}_t$ the domain in the space $\mathbb{R}^{n_u}$ where the admissible control values are defined. Backward recursion of the following stochastic continuous-time dynamic programming equations can give the optimal control of the above problem:
\begin{equation}\label{eq:dp1}
J_{t-1}(\Gamma_{t-1})=\underset{u(t-1)\in\Omega_{t-1}}{\min}\bigg[\frac{1}{t}\int_0^t\Bar{e}^2(\tau)d\tau|\Gamma_{t-1} \bigg],
\end{equation}
\begin{equation}\label{eq:dp2}
\begin{split}
&J_t(\Gamma_t)=\underset{u(t)\in\Omega_t}{\min}\bigg[\frac{1}{t}\int_0^t\Bar{e}^2(\tau)d\tau+J_{t+1}(\Gamma_{t+1})|\Gamma_t\bigg],\\
&\text{for }t=t-2,t-3,\dots,0.
\end{split}
\end{equation}

This optimal solution is usually impractical to derive analytically from \eqref{eq:dp1} and \eqref{eq:dp2} because of the dimension of the underlying spaces, even for simple cases. Near-optimal solutions can be obtained with model-based adaptive methods.

\subsection{Model-based adaptive control}
The design of a controller usually begins with a mathematical model of the system to be controlled. For mechanical systems of moderate dimension, it is possible to write down such a model (e.g. based on the Newtonian, Lagrangian, or Hamiltonian formalism) and eventually linearize its dynamics around a fixed point or periodic orbit. Model-based adaptive control refers to adaptive control methods that are completely reliant on this type of model. This means that the feedback control law will adjust the controller parameters online to compensate for model uncertainty. This adaptation can be performed in different manners, which gives the classification of model-based methods in two groups: direct and indirect.
In direct schemes, designers attempt to estimate the control parameters. The adjustment rule tells directly how the controller's parameters should be updated. It is done without intermediate calculations involving plant parameter estimates. This is possible because in many cases, there exist measurable variables that correlate well with changes in the process dynamics. The \textit{Gain Scheduling}~\cite{Stein1980ADAPTIVEFC} and \textit{Model-Reference Adaptive Systems}~\cite{WhitakerMRAS} algorithms are examples of direct schemes. On the other hand, if the estimates of the process parameters are updated and the controller parameters are obtained from the solution of a design problem using the estimated parameters, we obtain what is called an indirect approach. This kind of controller can be seen as automation of process modeling and design, in which the process model and the control design are updated at each sampling period. The \textit{Self-Tuning Regulators}~\cite{BorissonSTR} and \textit{Dual Controllers}~\cite{SternbyDual} are well-known indirect methods.
Instead of further extending the theory basis of model-based control (since it is not the main focus of this study), we describe here recent applications to the control of quadrotor MAVs under wind disturbances. An output controller was proposed in~\cite{Pyrkin2014OutputCF} to cancel wind disturbances where the nonlinear dynamical mathematical model of a quadcopter was decomposed into two parts: a static MIMO transformation and few SISO channels.  It was assumed that the unknown wind force acts on each channel of the quadcopter as a constant signal that has to be canceled. They proposed to first design a virtual control law for each SISO channel and then to inverse the MIMO transformation to get the control laws for the initial considered system. Their approach showed, under simulation, that it was able to efficiently reach steady-state under unknown wind perturbations. Later on, to reduce the undesirable effects of wind on a similar quadcopter, A. Razinkova et al.~\cite{Razinkova2014AdaptiveCO} chose another strategy. They assumed the wind to be an unknown but uniform and time-varying force acting on the vehicle body in the $X$ and $Y$-axes. Therefore, they proposed to augment a PD controller with additional terms adapting to the aforesaid forces. The resulting PD control law according to an axis $i$ takes the form: $u_i=K_p\times e(t)+K_d\times e(t)/d_t-\gamma\tilde{i}$, where $e(t)$ is the position error and $\gamma\tilde{i}$ the adaption term. Such an adaptation converges~\cite{Razinkova2014AdaptiveCO} if and only if the adaptation rate $\gamma$ remains positive. Later on, in~\cite{Fernndez2017L1AC} a $\mathcal{L}_1$ adaptive controller~\cite{Cao2007L1AO,Cao2008adaptiveCF} (which is an advanced version of MRAC~\cite{WhitakerMRAS}) was proposed to control a quadcopter that is performing wind turbine inspection. They proved the robustness of their controller with respect to wind under simulated and real flights by comparing it to a basic Linear Quadratic Regulator (LQR). More recently, in~\cite{Kim20202OP} the gains of a PID controller have been tuned especially in order to handle wind gusts. They tuned this classical feedback controller in the $\mathcal{H}_2$ optimal control framework and compared it with the existing LQR-based tuning method~\cite{Argentim2013PIDLA}. Simulated experiments proved that the classical PID controller combined with a dedicated parameter adjustment mechanism is better at rejecting wind disturbances than the classical LQR method.

These approaches have in common the use of the \textit{Certainty Equivalence} (CE) approach, which means that the estimation uncertainty is not taken into consideration. The wind estimates are used in the control law as if they were the real values of the unknown wind field. The CE paradigm has been used for a long time and makes it possible to generate a wide class of model-based adaptive controllers by combining different on-line parameter estimation algorithms (e.g. sensitivity methods, positivity and Lyapunov design, gradient and least-squares methods, etc) with different control laws (e.g. PID, LQG, LQR, Pole placement, etc). However, it was proved in~\cite{FilatovDual} that control systems based on the CE approach are not always optimal and can be far from so. A straightforward improvement would be to apply reinforcement on the uncertainty estimates over time, in a sort of \textit{learning process}, as a mean to optimize control performance. This goal, combined with the development of complex systems, where precise modelization is extremely difficult, has motivated the development of the model-free adaptive control field.

\subsection{Model-free adaptive control} \label{model_free}
\begin{figure}[H]
   \begin{center}
      \includegraphics[scale=0.5]{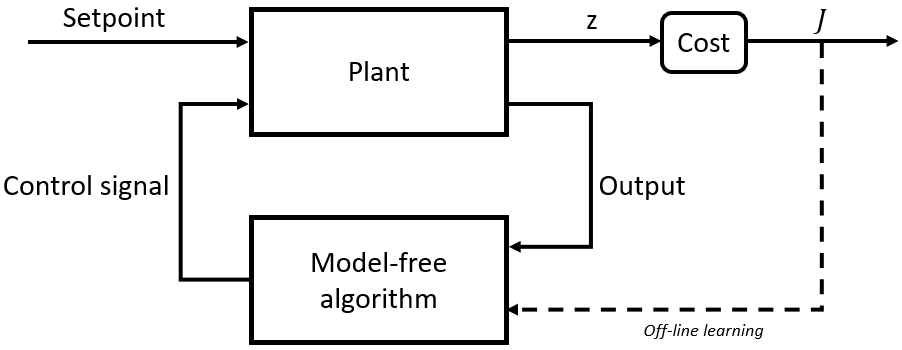}
   \end{center}
   \caption[todo]
   {\footnotesize General block diagram of a classical model-free controller. The control objective is to minimize a well-defined cost function $J$ within the space of possible control laws. The off-line learning loop provides experiential data to train the controller. The vector $z$ is composed of all the information that may factor into the cost.}
   \label{fig:modelfreebloc}
\end{figure}

The \textit{model-free} wording refers here to the fact that these families of controllers do not rely on any mathematical model of the controlled system. They aim at describing complex systems from observational data collected directly by embedded sensors rather than first-principles modeling. As shown in Figure \ref{fig:modelfreebloc}, model-free algorithms can be seen as an optimization problem where the goal is to minimize a cost function without closed-form knowledge of the function or its gradient. A classic approach widely used in the model-free control framework is the \textit{Extremum-Seeking} (ES) methods~\cite{AriyurES,Zhang2011ExtremumSeekingCA}. We present now a simple ES algorithm by considering the same equations of state \eqref{adaptproblem}. The goal of the control is again to optimize the performance of the system described by the cost function~\eqref{perfindex}. We model it as a smooth function $J(x,u):\mathbb{R}^n\times\mathbb{R}\rightarrow\mathbb{R}$. We will denote it simply by $J(u)$ because the state vector $x$ is driven by $u$. 
In order to write convergence results we need the following assumptions:
\begin{enumerate}
    \item There exists a smooth function $l:\mathbb{R}\rightarrow\mathbb{R}^n$ such that:
    \begin{equation}
        f(x(t),u(t),p(t),t)=0, \text{if and only if }x(t)=l(u(t))
    \end{equation}
    \item For each $u\in\mathbb{R}$, the equilibrium $x=l(u)$ is locally exponentially stable.
    \item There exists (a maximum) $u*\in\mathbb{R}$ such that:
    \begin{equation}
    \begin{split}
        &(J \circ l)^{(1)}(u*)=0 \\
        &(J \circ l)^{(2)}(u*)<0
    \end{split}
    \end{equation}
\end{enumerate}

\noindent Based on these assumptions, one can easily design a simple extremum seeker with proven convergence bounds. One of the simplest way to maximize $J$ is to use a gradient-based ES control law as:
\begin{equation}\label{eq:esalgo}
    \dot{u}=k\frac{dJ}{du},\hspace*{5px}k>0
\end{equation}

\noindent The convergence of \eqref{eq:esalgo} can be analysed with the Lyapunov function:
\begin{equation}
    V=J(u*)-J(u)>0,\hspace*{5px}\text{for $u\neq u*$}
\end{equation}

\noindent The derivative of $V$ gives:
\begin{equation}
    \dot{V}=\frac{dJ}{du}\dot{u}=-k\left(\frac{dJ}{du}\right)^2\leq 0
\end{equation}

This proves that the algorithm drives $u$ to the invariant set $\frac{dJ}{du}=0$, which is equivalent to $u=u*$. However, as simple as it seems, this approach requires the knowledge of the gradient of $J$. An ES-based model-free approach was proposed in~\cite{Tagliabue2019ModelfreeOM} for the optimal control of a quadcopter carrying different payloads. They proved, with experimental flights, that their model-free controller is able to find the speed that maximizes the flight time (endurance) or flight distance (range) of the vehicle when transporting an unknown payload. They designed specific cost functions for each of these goals. Then, the ES algorithm was applied to estimate an optimal velocity which is used to transform the desired path into a trajectory. They proved that the ES scheme is able to find unknown, time-varying, operating points that minimize these cost functions directly from sensorial measurements and without any model of the MAV power consumption.

Model-free control is a blossoming field where a handful of strong techniques are currently being developed and applied to minimize or maximize specific cost functions by using the system outputs as shown in Figure~\ref{fig:modelfreebloc}. Reinforcement learning (RL) is currently the leading technique used in this research area. It is a class of machine learning methods in which an autonomous agent in an environment has to learn how to take actions in order to maximize the notion of cumulative reward~\cite{Sutton2005ReinforcementLA}. More formally, the objective of RL is for a learning agent to find an optimal strategy behavior (called a policy and denoted by $\pi$) from experimental trials and errors. Those repeated interactions are under the form of the execution of an action $a_t\in A$ from state $s_t\in S$ which makes the agent transit to a new state $s_{t+1}\in S$. This transition produces a reward signal
denoted by $r(s_t,a_t)\in \mathbb{R}$, which quantitatively transcribes how well the agent is doing in the environment. The policy is a rule used by the agent to decide what actions to take and is mathematically defined as a function:
\begin{equation}
\pi:S \rightarrow A
\end{equation}

This type of procedure may be framed as a Markov Decision Process~\cite{Howard1960} along a trajectory (a sequence of $ T $ actions and $ T+1 $ states in the environment) $\tau = (s_0,a_0,s_1,a_1,\dots,a_{T-1},s_{T})$. Hence, the probability of a trajectory for a $\pi$ policy is:
\begin{align}
	P\left(\tau|\pi\right)=\rho_0(s_0) \prod\limits_{t=0}^{T-1} P(s_{t+1}|s_t,a_t)\pi(a_t|s_t)
\end{align}
where $\rho_0(s_0) $ means that the starting state $ s_0 $ is randomly sampled from the distribution $ \rho_0 $ and $ \pi(a_t|s_t) $ means that the action $ a_t $ is randomly sampled by the policy $ \pi $ in the state $ s_t $. The overall optimization problem considered in RL is to determine the policy $ \pi^\star $ which maximizes the expected return when the agent decides to take actions according to this policy:
\begin{equation}  \label{RLproblem}
	\pi_{\star} = \argmax J_{RL}(\pi) 
\end{equation}
where $ J_{RL}(\pi)  $ denotes the expected return
\begin{equation} \label{JRL}
J_{RL}(\pi) = \int_{\tau} \sum\limits_{t=0}^{T-1}r_t \cdot P\left(\tau|\pi\right)  =\underset{\tau\sim\pi}{E}\left[\sum\limits_{t=0}^{T-1}r_t\right] 
\end{equation}

Both RL and model-free paradigms are based on the same optimization formalism. We can easily see the similarity between this cost function \eqref{JRL} and the one from the ES algorithm \eqref{eq:esalgo}. Note that the terms model-based and model-free can also be found in the RL theory but do not correspond to the ones used in the control field. In RL, they refer to whether or not a model of the environment (a function which predicts state transitions and rewards) is given beforehand to the RL algorithm to improve the policy whereas, in control, it is the model of the controlled dynamical system that might be provided.

Various methods exist to tackle the RL optimization problem. Among them, \textit{Policy Gradient} techniques have enabled to use RL in real-world robotic contexts (see~\cite{Schulman2016} for an extensive definition of these approaches and~\cite{Loquercio2018DroNetLT,Kahn2018SelfSupervisedDR,Soans2020SANetRS} for successful real-world robotic applications). These techniques are based on the use of a stochastic policy formally denoted by $\pi_\theta$, where $\theta$ is a vector of parameters. The vector $\theta$ is estimated iteratively by gradient ascent.
The expression of the gradient of $J(\pi_\theta)$ with respect to $\theta$ is given by:
 \begin{align}
     \nabla_\theta J_{RL}(\pi_\theta) = \underset{\tau\sim\pi_\theta}{E}\left[\sum\limits_{t=0}^{T} \nabla_\theta \log \pi_\theta(a_t|s_t)  (Q^{\pi_\theta}(s_t,a_t)-V^{\pi_\theta}(s_t)) \right]
 \end{align}
where $V^{\pi_\theta}(s_t)$ is $V$-value function quantifying the expected return if the trajectory starts in a state $s_t$ taking actions upon the policy $\pi_{\theta}$:
 \begin{align}
 V^{\pi_\theta}(s_t) = \underset{\tau\sim\pi_\theta}{E} \left[ \sum\limits_{t=0}^{T} r_t |s_0=s_t \right]
 \end{align}
 and $ Q^{\pi_\theta}(s_t,a_t)$ is the $Q$-value function quantifying the expected return if the trajectory starts in a state $s_t$, takes an action $a_t$ and then takes actions upon the policy 
$\pi_{\theta}$:
 \begin{align}
 Q^{\pi_\theta}(s_t,a_t) = \underset{\tau\sim\pi_\theta}{E} \left[ \sum\limits_{t=0}^{T} r_t |(s_0=s_t,a_0=a_t) \right]
 \end{align}

\subsection{Learning-based adaptive control} \label{learning_based}
Learning-based controllers can be seen as hybrid control schemes. They are used when we only have access to an imperfect physics-based model of the system. The idea is to strengthen it with some model-free algorithm to compensate for the uncertain or the missing parts of the model. This compensation can either be done directly by \textit{learning} the uncertain parts or indirectly by \textit{tuning} the controller parameters to cope with the uncertainty. To illustrate this idea, we consider the system \eqref{adaptproblem} with a specific structure that can be written as:
\begin{equation} \label{model:learning-based}
\begin{split}
&\dot{x}(t)=f_1(x(t),u(t),t)+f_2(x(t),p(t),t),\\
&y(t)=h(x(t),u(t), t)
\end{split}
\end{equation}

In this model, $f_1$ represents the known part of the plant that can be efficiently driven by a classical model-based control architecture, while $f_2$ represents the unknown part of the model that can be compensated by some model-free learning algorithm. The goal of learning-based controllers is to use a model-free step to optimize an unknown performance function and then use a model-based control law to guide the system's dynamics toward the optimal performance.
The idea of combining both techniques is indeed attractive. One designer could take advantage of the model-based design, with its stability characteristics, and add to it the advantages of model-free learning, with its fast convergence and robustness to uncertainties. Overall, in Neural-Network (NN) learning-based control design, the idea is to write the model of the plant as a combination of a known and unknown part (i.e. the disturbances). The NNs are then used to estimate the unknown part of the model. As a result, a controller based on the known part (model-based) and the NN estimates of the uncertainties (model-free) is determined to realize some desired regulation or tracking performance. To illustrate this concept we formalize it below, in a similar spirit as the formalism from~\cite{Mouhacine16} where a state-space model of \eqref{adaptproblem} under the Brunovsky form is considered as:
\begin{equation} \label{eq:brunovsky}
\left\{
\begin{array}{l}
\dot{x}_1=x_2,\\
\dot{x}_2=f(\cdot)+u,\\
\end{array}
\right.
\end{equation}
where $f(\cdot)=f_1(x(t),u(t),t)+f_2(x(t),p(t),t)$, with $f_1$ known and $f_2$ an unknown smooth function of the state variables $x=(x_1, x_2)^T$ and $u$ the control signal. The unknown part of the model, namely $f_2$, is estimated by a NN as:
\begin{equation}
    \hat{f}_2=\hat{W}^TS(x(t)),
\end{equation}
where $\hat{W}=(\hat{w}_1,\dots,\hat{w}_N)^T\in\mathbb{R}^N$ is the estimated vector of synapse weights of the neural network node and $S(x)=(s_1(x),\dots,s_n(x))^T$ is the regressor vector, with \mbox{$s_i,i=1,\dots,N$}. Consider the reference model:
\begin{equation}
\left\{
\begin{array}{l}
\dot{x}_{ref_1}=x_{ref_2},\\
\dot{x}_{ref_2}=f_{ref}(x),\\
\end{array}
\right.
\end{equation}
where $f_{ref}$ is a known nonlinear smooth function of the desired trajectories $x_{ref}=(x_{ref1}, x_{ref2})^T$. A basic learning-based controller can be defined as:
\begin{equation} \label{NNs-control-law}
    u = -e_1-c_1e_2 -\hat{W}^TS(e) +\dot{v},
\end{equation}
with:
\begin{equation}
    \begin{split}
        &e_1=x_1-x_{ref1},\\
        &e_2=x_2-v,\\
        &v= -c_2e_1+x_{ref2}\\
        &\dot{v}=-c_2(-c_2e_1+e_2)+f_{ref}(x_{ref}),\hspace{5px}(c_1,c_2>0)\\
        &\dot{\hat{W}}=\Gamma(S(e)e_2-\sigma\hat{W}),\hspace{5px}(\sigma>0\text{ and }\Gamma^T>0)
    \end{split}
\end{equation}

It can be observed that $u$ is now a function of the Brunovsky form  \eqref{eq:brunovsky} (i.e.,~model-based information), and the remaining part is based on the NN estimates of $f_2$ (model-free estimation). The unknown part is here estimated by a neural network but other strategies can be used in the learning-based scheme.

This concept of using NNs to estimate the unknown parameters (i.e. the wind field for our use case) seems appealing. In fact, as stated by the universal approximation theorem~\cite{LewisNNs} for any function $f(x)$, regardless of its complexity, there exists a neural network such that for every possible input $x$, the value $f(x)$ is a feasible output from this network. If the external disturbances are well estimated, efficient control laws can be designed accordingly. This idea was investigated in~\cite{Bisheban2018GeometricAC,Bisheban2019GeometricAC} where a \textit{Geometric Adaptive Controller} was proposed based on NNs for a quadcopter in wind fields. Two control laws were defined with adaptive control terms denoted as $\Bar{\nabla}_i$ to mitigate the effects of the unknown disturbance. These adaptive control terms were computed with neural networks, exactly like in  \eqref{NNs-control-law}, as: $\Bar{\nabla}_i=\Bar{W}^T_i\varsigma(\Bar{z}_i)$ with $\Bar{z}_i=\Bar{V}^T_ix_{nn_i}$ ($\Bar{V}_i$ being the current estimate of the ideal weighting parameters and $x_{nn_i}$ the inputs of the neural network). They proved with real experiments that the learning-based controller succeeded to complete the considered backflip maneuver followed by a stable hovering flight whereas it was not possible without the disturbance rejection terms. Another strategy was proposed in~\cite{Zheng2020SafeOL}, where a learning-based safety-preserving
cascaded QP controller (SPQC) using \textit{Gaussian Processes} (GP)~\cite{Rasmussen2010GaussianPF} has been proposed for safe trajectory tracking by a quadcopter in a cluttered environment. More precisely, the cascade controller is composed of two QP controllers: a position and an attitude level QP controller. The first one generates the desired thrust while the second makes use of it, together with the high confidence uncertainty interval obtained via GPs, to compute the desired body rotational rates. They evaluated their approach under numerical simulation whose results proved that the proposed learning-based controller is able to perform a trajectory tracking task with obstacle avoidance capacity under changing wind fields.

Compared to the approaches presented in this related work Section, we propose to only treat the parameters adjustment task. We design a learning-based controller that uses a DRL-based model-free algorithm to perform online-tuning of the parameters of a state-feedback controller. We compare this method to the model-free strategy initially applied in~\cite{TChaffre} but this time for the application of a waypoint rallying mission by a MAV under unknown wind gusts.


\section{MAV simulation, modeling and control} \label{background}
This Section presents the ROS package used to simulate the aerial vehicle and wind perturbations and the corresponding model of the Firefly platform, which is the hexacopter considered in this study. The principal elements necessary to understand the control designs proposed in Section~\ref{method} are also derived.

\subsection{RotorS package} \label{RotorS}
Testing algorithms on physical platforms can be very time-consuming and dangerous for the robot. It can be even riskier when one designer wants to perform the training of machine learning algorithms directly on the real robot (because of their initial hazardous behavior). There exist many robotic simulation tools to reduce field testing time and make debugging easier. In this work, we relied on the ubiquitous Gazebo simulator~\cite{Koenig2004DesignAU} which is connected to the Robot Operating System framework (ROS~\cite{Quigley2009ROSAO}). ROS is an open-source meta-operating system for robots that facilitates the reuse of code.

\begin{figure}[!h]
\begin{subfigure}{0.5\textwidth}
\centering
\includegraphics[width=\textwidth]{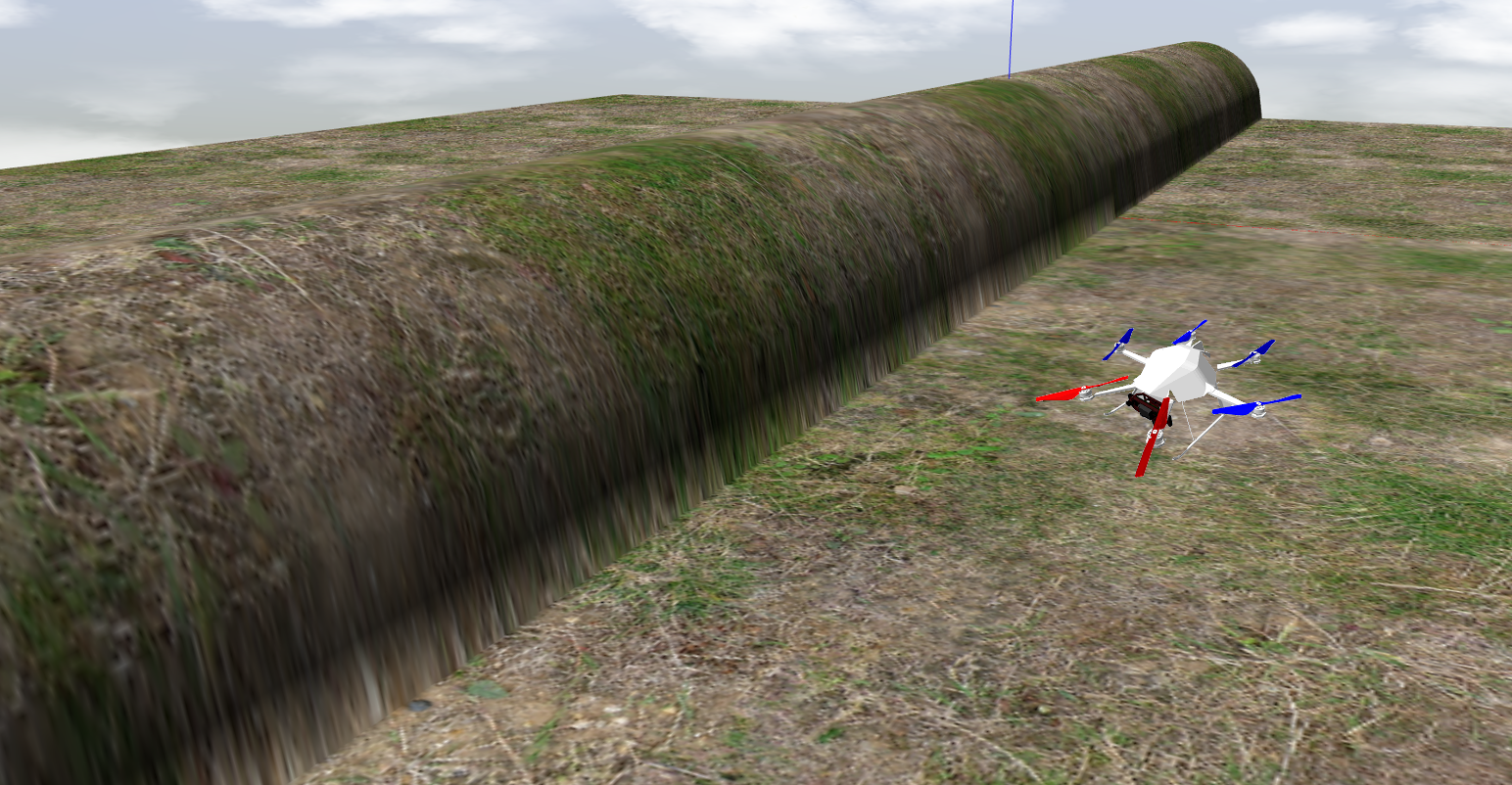} 
\caption{}
\label{fig:gazebo}
\end{subfigure}
\begin{subfigure}{0.5\textwidth}
\centering
\includegraphics[width=\textwidth]{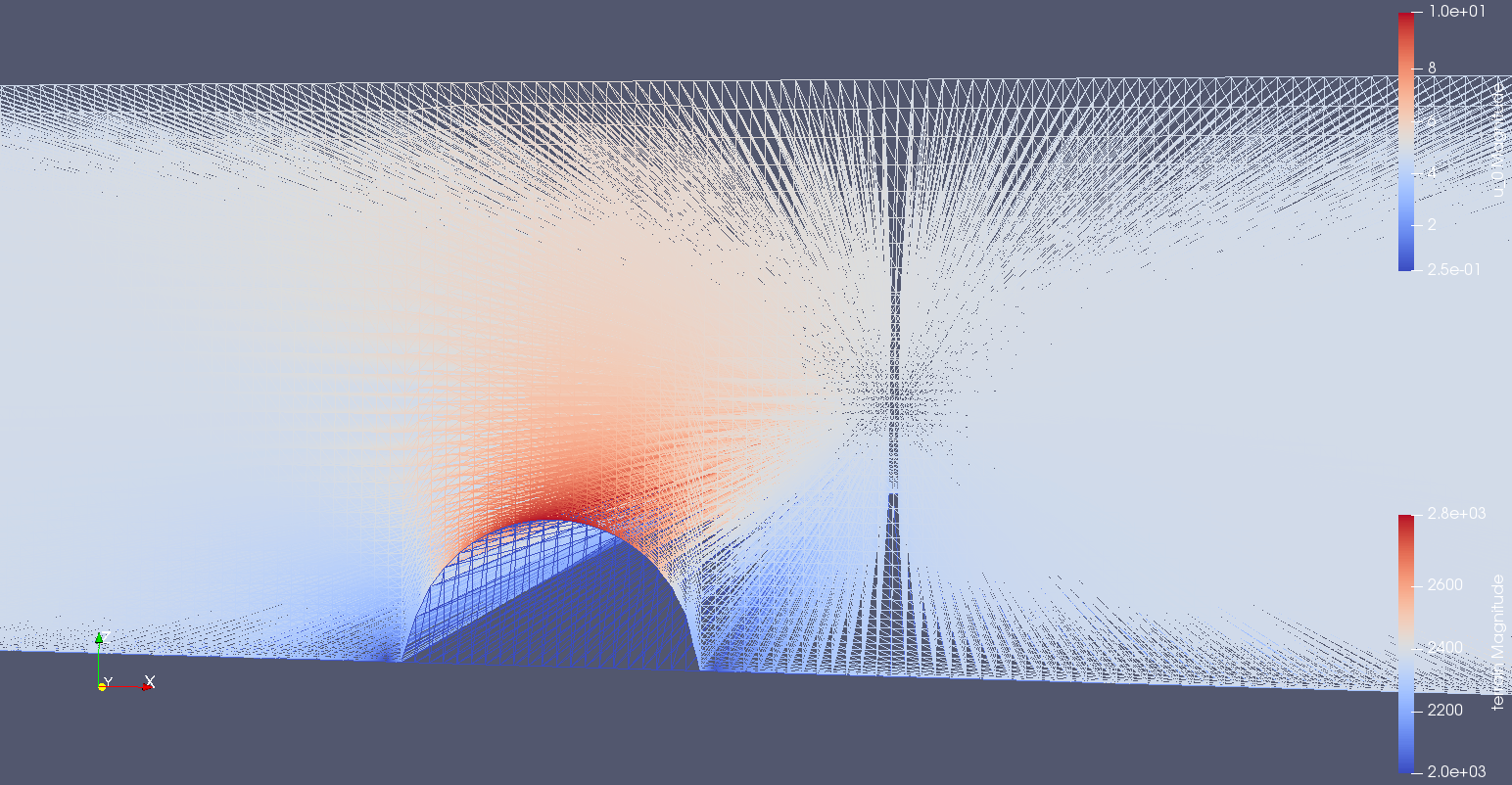}
\caption{}
\label{fig:paraview}
\end{subfigure}
\caption{Visualization of the simulated environment in Gazebo (a) and the wind field in Paraview (b). The complete details of the wind field we used can be find on the RotorS-based \href{https://github.com/ethz-asl/rotors_simulator/wiki/Adding-a-custom-wind-field-to-your-world}{extended wind plugin} page.}
\label{environment}
\end{figure}

The Gazebo-based package called RotorS~\cite{Furrer2016} is a high-fidelity simulation framework for MAVs (developed by the Autonomous Systems Lab team from ETH Zurich) that does not require any additional components to simulate high-level tasks (e.g. path-planning, collision avoidance, or vision-based problems). A complete model of the Firefly MAV is directly included in the package along with various world models and a plugin to simulate wind fields. We relied on the wind plugin provided by RotorS to generate wind fields in the environment. This plugin allows to define the wind as a 3D field sampled over a regular grid and each point specifies a wind velocity vector (in $m.s^{-1}$). More specifically, we employed the environment named \textit{hemicyl} (see Figure \ref{environment}) and its pre-configured wind field, the field used is composed of $6282$ vertices and the wind velocity vary here between $-5m.s^{-1}$ and $+10m.s^{-1}$.

\newpage
\subsection{MAV model} \label{mav_model}

The parametrization of the Firefly platform considers two reference frames: the world-fixed inertial frame $\mathcal{R}_W$ and the body reference frame $\mathcal{R}_B$ which is attached to the center of mass of the hexacopter. Coordinates in the world frame are denoted as $[x_W,y_W,z_W]^T$ while they are denoted as $[x_B,y_B,z_B]^T$ in the body frame. The pose of the hexacopter is given by its position $\zeta=[x_W,y_W,z_W]^T$ and orientation $ \eta=[\phi,\theta,\psi]^T$ in the three Euler angles (respectively roll, pitch and yaw). For the sake of clarity, $\sin(\cdot)$ and $\cos(\cdot)$ are abbreviated as $s\cdot$ and $c\cdot$ in the next equation. The transformation from the world frame $\mathcal{R}_W$ to the body frame $\mathcal{R}_B$ is given by:
\begin{equation}
\begin{bmatrix}
x_B \\
y_B \\
z_B\end{bmatrix}=
\begin{bmatrix}
c\theta c\psi &\hspace{10px}c\theta s\psi &\hspace{10px}-s\theta \\
s\phi s\theta c\psi - c\phi s\psi &\hspace{10px}s\phi s\theta s\psi + c\phi c\psi &\hspace{10px}s\phi c\theta\\
c\phi s\theta c\psi + s\phi s\psi &\hspace{10px}c\theta s\theta s\psi - s\phi c\psi &\hspace{10px}c\phi c\theta
\end{bmatrix}
\begin{bmatrix}
x_W \\
y_W \\
z_W\end{bmatrix}
\end{equation}

The main forces acting on the vehicle come from gravity and the thrust of the rotors. Rotors drag and air friction are neglected here to simplify the model. 
It is classically assumed that the hexacopter is a rigid body with a symmetrical structure, and tensions in all directions are proportional to the square of the propeller speed. The equations of motion follow as:
\begin{equation}
\begin{split}
&\dot{\zeta}=v\\
&\dot{v}=-g e_3 +\textbf{R}\left(\frac{b}{m}\sum\Omega^2_i\right)\\
&\dot{\textbf{R}}=\textbf{R}\hat{\omega}\\
&\textbf{I}\dot{\omega}=-\omega\times \textbf{I}\omega-\sum \textbf{$J_r$}(\omega\times e_3)\Omega_i\tau 
\end{split}
\end{equation}
where \textbf{R} the rotation matrix from $\mathcal{R}_B$ to $\mathcal{R}_W$~
; ${\omega}$ is the skew symmetric matrix~; $\Omega_i$ is the $i$-th rotor speed ; \textbf{I} the body inertia ; \textbf{$J_r$} the rotor inertia ; $b$ is the thrust factor and $\tau$ is the torque applied to the body frame due to the rotors. 
\noindent A classical cascade control structure is adopted~\cite{bertrand2011hierarchical}, where the built-in low-level controller of the RotorS package is used to track a reference in roll~$\phi_{\mathrm{r}}$, pitch~$\theta_{\mathrm{r}}$ and thrust~$\mathcal{T}$. The yaw angle $\psi$ is kept constant without loss of generality. Under the small-angle assumption on $\phi$ and $\theta$, the guidance model reduces to the double-integrator model:
\begin{equation}
    \begin{split}
&\dot{\zeta}=v\\
&\dot{v}=u = [u_x,u_y,u_z]^T
\label{doubleint}
\end{split}
\end{equation}
where the computed accelerations are converted into low-level control inputs as: \begin{equation}
\begin{split}
&\mathcal{T} = m(u_z + g) \\
&\theta_\mathrm{r} = \frac{m}{\mathcal{T}}(c\psi u_x + s\psi u_y) \\
&\phi_\mathrm{r} = \frac{m}{\mathcal{T}}(s\psi u_x - c\psi u_y)
\end{split}\label{controlinputs}
\end{equation}


\subsection{Controller parametrization} \label{polePlacement}
For future use by the learning-based control approach (Section~\ref{lbcontroller}), a specific parametrization is adopted to control the system~\eqref{doubleint}.
The control objective is to stabilize the robot at a waypoint $\zeta_\mathrm{ref}$ with a velocity $v=0$. Defining $e = \zeta-\zeta_{\mathrm{ref}}$ 
and taking into account unmodeled disturbances (e.g. wind) by considering an additional steady-state error variable ${z=\int_{0}^{t}e(\tau)d\tau}$, the augmented model with state vector $X=\left[z,e,v\right]$
becomes:
\begin{equation}
\left[\begin{array}{c}
\dot{z}\\
\dot{e}\\
\dot{v}
\end{array}\right]=\left[\begin{array}{ccc}
0 & 1 & 0\\
0 & 0 & 1\\
0 & 0 & 0
\end{array}\right]\left[\begin{array}{c}
z\\
e\\
v
\end{array}\right]+\left[\begin{array}{c}
0\\
0\\
1
\end{array}\right]u
\end{equation}
The corresponding state-feedback controller, equivalent to a PID, is such that:
\begin{equation} \label{controlLAW}
u=-k_{i}z - k_{p}e - k_{d}v
\end{equation}
The poles of the closed-loop system are solution to the following equation:
\begin{equation}
    \lambda^{3}+\lambda^{2}k_{d}+\lambda k_{p}+k_{i}=0 \label{polcar}
\end{equation}

The controller has been re-parametrized using pole placement so as to guarantee convergence to steady-state without oscillations. This way, the action space for learning purposes is limited to desired solutions in the real part of the pole map, which prevents from sampling unnecessary solutions in the space of the control gains. The desired constants $\tau_{1}>0,\tau_{2}>0,\tau_{3}>0$ are then defined as:
\begin{equation}
\lambda_{1}=\frac{-1}{\tau_{1}}\,;\,\lambda_{2}=\frac{-1}{\tau_{2}}\,;\,\lambda_{3}=-\frac{1}{\tau_{3}}
\end{equation}
Since each one is solution to~\eqref{polcar}, it follows that:
\begin{gather}
\left[\begin{array}{ccc}
1 & \frac{-1}{\tau_{1}} & \frac{1}{\tau_{1}^{2}}\\
1 & \frac{-1}{\tau_{2}} & \frac{1}{\tau_{2}^{2}}\\
1 & \frac{-1}{\tau_{3}} & \frac{1}{\tau_{3}^{2}}
\end{array}\right]\left[\begin{array}{c}
k_{i}\\
k_{p}\\
k_{d}
\end{array}\right]=\left[\begin{array}{c}
\frac{1}{\tau_{1}^{3}}\\
\frac{1}{\tau_{2}^{3}}\\
\frac{1}{\tau_{3}^{3}}
\end{array}\right]\Leftrightarrow MK^{T}=N
\end{gather}
Finally, the gains of the controller~\eqref{controlLAW} are obtained as $K^{T}=M^{-1}N$:
\begin{equation}
\begin{split}
&k_{i}=\frac{1}{\tau_1\tau_2\tau_3}\\ \\
&k_{p}=\frac{\tau_1+\tau_2+\tau_3}{\tau_1\tau_2\tau_3}\\ \\
&k_{d}=\frac{\tau_1\tau_2+\tau_1\tau_3+\tau_2\tau_3}{\tau_1\tau_2\tau_3}
\end{split}
\label{control_gains}
\end{equation} 

\newpage
\section{Methodology} \label{method}
In this Section, we present how to build a model-free and learning-based controller with a DRL framework composed of two components: the Soft Actor-Critic algorithm and the Experience Replay technique. Later, we present how to apply it to the model-free and learning-based control schemes.
\subsection{Soft Actor-Critic}\label{SAC}
In this study, we used the same policy gradient algorithm as in our previous paper~\cite{TChaffre} to either maximize the RL-based cost function of a model-free controller (see Section \ref{model_free}) or to estimate the controller parameters of a learning-based controller (as described in Section \ref{learning_based}). The \textit{Soft Actor-Critic} (SAC) algorithm proposed by T. Haarnoja et al. in~\cite{haarnojaSoft} is based on three concepts which we formally define thereafter.

First, the SAC algorithm uses an actor-critic architecture which concurrently learns both State value and Q-value functions and the policy $\pi_\theta(a_t|s_t)$ as well. During the training stage, the critic part updates parameters of the $State$ or $Q$-value functions while the actor network updates the policy parameters in the direction suggested by the critic. Note that a recent version of the SAC algorithm~\cite{Haarnoja2018SoftAA} allows to only estimate the Q-value function, while we used in this work the original version of the algorithm~\cite{haarnojaSoft} where the State value function is also estimated to help stabilizing the overall training process. Precisely, our implementation of the SAC algorithm aims at iteratively learning three functions modelized by three NNs: 
\begin{itemize}
    \item a policy function $\pi_\theta(a_t|s_t)$ parameterized by the NN weights $\theta$,
    \item a soft Q-value function $Q_w(s_t,a_t)$ parameterized by the NN weights $w$,
    \item and a soft State value function $V_\Psi(s_t)$ parameterized by the NN weights $\Psi$.
\end{itemize}

Beyond their modelization, these soft Q-value and State value functions are induced by the policy $\pi_\theta$ and defined as follows:
\begin{equation} \label{targetQ}
Q_{w}^{\pi_\theta}(s_t,a_t)=r_t(s_t,a_t)+\gamma\mathbb{E}_{s_{t+1}\sim\rho_\pi(s)}[V_{\Psi}^{\pi_\theta}(s_{t+1})]
\end{equation}
\begin{equation} \label{targetV}
V_{\Psi}^{\pi_\theta}(s_t)=\mathbb{E}_{a_t\sim\pi_{\theta}}[Q_{w}^{\pi_\theta}(s_t,a_t)-\alpha\log{\pi_{\theta}(a_t|s_t)}]
\end{equation}

The terms $\rho_\pi(s)$ and $\rho_\pi(s,a)$ denote the state and the state-action marginals of the state distribution induced by the policy $\pi_\theta(a|s)$. The NNs parameters $w$ and $\Psi$ of the Q-value and State value functions respectively, are the solutions minimizing the soft Bellman residual $J_Q(w)$ and the mean squared error $J_V(\Psi)$:
\begin{equation}\label{JQ}
\begin{split}
J_Q(w)=\mathbb{E}_{(s_t,a_t)\sim D}\Bigg[\frac{1}{2}\bigg(Q_w^{\pi_\theta}(s_t,a_t)&-\Big(r(s_t,a_t)\\ &+\gamma\mathbb{E}_{s_{t+1}\sim\rho_\pi(s)}[V_{\Bar{\Psi}}^{\pi_\theta}(s_{t+1})]\Big)\bigg)^2\Bigg]
\end{split}
\end{equation}
\begin{equation}\label{JV}
J_V(\Psi)=\mathbb{E}_{s_t\sim D}\left[\frac{1}{2}(V_\Psi(s_t)-\mathbb{E}[Q_w^{\pi_\theta}(s_t,a_t) -\log{\pi_\theta}(a_t,s_t)])^2\right]
\end{equation}
with $\Bar{\Psi}$ the target value function. The NN parameters $\theta$ of the policy are updated in order to minimize the expected  Kullback-Leibler divergence:
\begin{equation}\label{Newpi}
    J_\pi(\theta)=\mathbb{E}_{s_t\sim D}\bigg[D_{KL}\bigg(\pi_\theta(\cdot|s_t) \bigg|\bigg|\frac{\exp(Q_w^{\pi_\theta}(s_t,\cdot))}{Z_w^{\pi_\theta}(s_t)} \bigg) \bigg]
\end{equation}
with $Z_w^{\pi_\theta}$ a partition function used to normalize the distribution. 
The ($\cdot)\sim D$ means that the expected values are computed using the pair $(s_t,a_t)$ and $(s_t)$ sampled from a replay buffer $D$ (see Section \ref{ER}) in which the agent experience $e_t=(s_t, a_t, r_t, s_{t+1})$ is stored at each time-step. Moreover, the authors use an exponentially moving average, with a smoothing constant \mbox{$\tau=5e^{-3}$}, to update the target value network weights $w$ (the parameters $\Psi$ are also update in the same way since they are directly related to $w$). Thus these weights are constrained to change slowly from one iteration to another. The name \textit{Soft Actor-Critic} is based on this soft update procedure.
Secondly, the SAC algorithm optimizes a stochastic policy in an off-policy manner~\cite{Degris2012OffPolicyA}. This means that the policy used to explore the environment is different from the one that is being evaluated and improved. 
Finally, the policy is trained to maximize simultaneously the expected return and its entropy, which is an amount of informative randomness:
\begin{equation} \label{RLentro}
J_{SAC}(\pi)=\sum_{t=1}^T\mathbb{E}_{(s_t,a_t)\sim \rho_{\pi}}[r(s_t,a_t)+\alpha H(\pi_\theta(.|s_t))]
\end{equation}
where:
\begin{equation}
H(\pi_\theta(.|s))=-\sum\limits_{a\in A}\pi_\theta(a)\log \pi_\theta(a|s)
\end{equation}
The term $H(\pi_\theta)$ is the entropy measure of policy $\pi_\theta$ and $\alpha$ is a fixed temperature parameter that determines the relative importance of the entropy term (an automatic adjustment mechanism for $\alpha$ was proposed in~\cite{Haarnoja2018SoftAA} by considering the constrained optimization problem of maximizing the expected return while satisfying a minimum entropy constraint but was not used in our implementation). The authors state that entropy maximization leads to policies that have better exploration capabilities. Note that this RL problem \eqref{RLentro} is different from the initial RL problem \eqref{RLproblem}. The resulting Q-value and V-value functions will therefore also include this entropy term. Two additional Q-functions are used to reduce positive bias in the policy improvement step, which is known to degrade performance of value based methods~\cite{Hasselt2010DoubleQ,Fujimoto2018AddressingFA}. They are trained independently and the minimum of the two Q-functions is used 
as proposed in~\cite{Fujimoto2018AddressingFA}.

\begin{figure}[!htbp]
\begin{subfigure}{0.5\textwidth}
\centering
\includegraphics[scale=0.625]{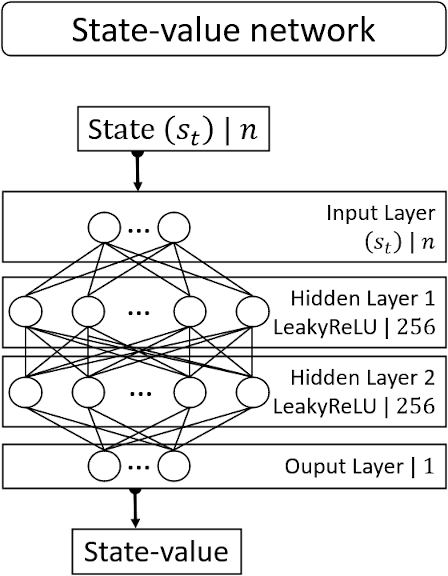} 
\label{fig:valuenet}
\end{subfigure}
\begin{subfigure}{0.5\textwidth}
\centering
\includegraphics[scale=0.625]{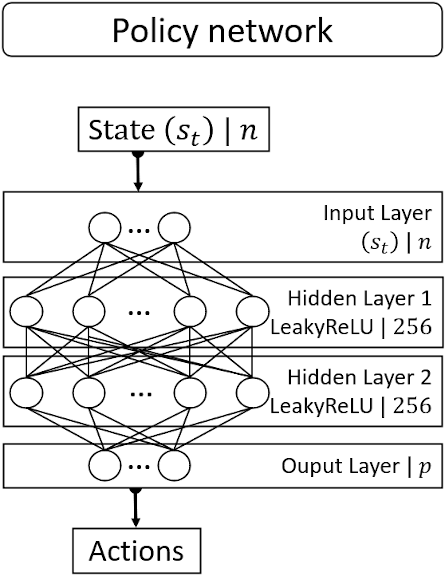}
\label{fig:policynet}
\end{subfigure}
\begin{center}
\begin{subfigure}{0.5\textwidth}
\centering
\includegraphics[scale=0.45]{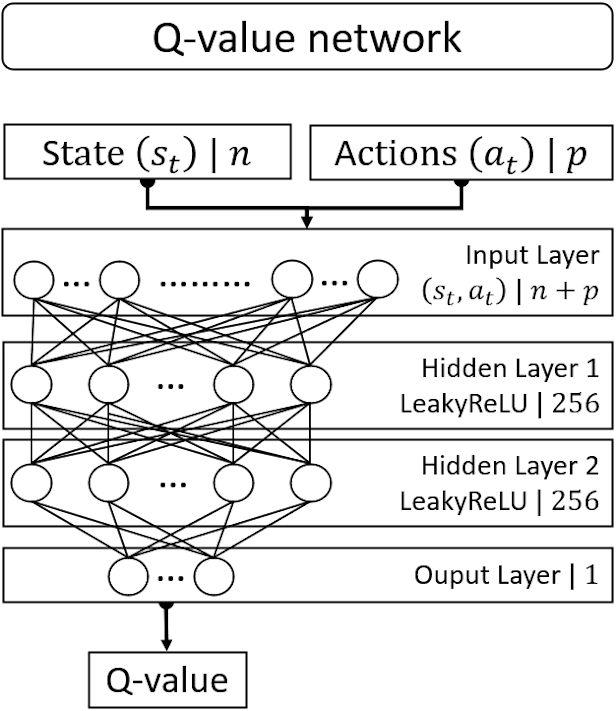}
\label{fig:qnet}
\end{subfigure}
\end{center}
\caption{The neural networks structure for our implementation of the SAC. The networks are composed of dense layers only and of two hidden layers for each network. Each layer is a fully-connected layer represented by its type, output size and activation function. The networks uses the same optimizer (Adam~\cite{DiederikAdam}), activation function (Leaky Relu~\cite{Xu2015EmpiricalEO}) and learning rate $l_r=3e^{-4}$. Other parameters are a discount factor $\gamma=0.99$ for \eqref{targetQ}, a number of $256$ hidden units per hidden layer and one gradient update is performed at each time-step.}
\label{fig:fignets}
\end{figure}

The SAC algorithm is part of the family of deep reinforcement learning approaches which try to exploit the strong representation capabilities offered by NNs to represent the Q-value, State-value and policy functions. The NNs structure of our SAC implementation is shown in Figure~\ref{fig:fignets}.

\newpage
\subsection{Experience Replay} \label{ER}
To improve its performance, one needs to learn from its past actions. In~\cite{LinER}, L.~Lin showed that past experience of the learning agent should and could be used in an effective way by using what he called \textit{Experience Replay} (ER). An \textit{experience} is the result of a transition to a new state which can be framed as a quadruplet $e_t=(s_t, a_t, r_t, s_{t+1})$. The ER technique consists in storing at each time-step the experience $e_t$ of the agent in a data-set (a replay buffer) $D=e_1,\dots,e_n$ of fixed and substantial size. During training, the RL algorithm updates are then applied to mini-batches of experiences randomly pooled over the stored samples to reinforce the functions estimates.

This intuitive concept can be refined by distinct means. It seems natural that when learning a new task, some experience might be more valuable than the others and should be used more often in the update process. This relevance is in general not directly accessible, therefore a specific criterion has to be chosen to quantify it. In~\cite{Schaul2016}, T. Schaul proposed the \textit{Prioritized Experience Replay} (PER) technique which consists in sampling more often from the replay buffer the transitions with high expected learning progress, as measured by the magnitude of their temporal-difference (TD) error. By doing so, they were able to obtain state-of-the-art performance on the Atari 2600 benchmark suite.

Nevertheless, depending on the task, PER can be very computationally expensive. Another parameter that has been considered is the replay capacity that is the total number of transitions possibly stored in the replay buffer. In~\cite{Zhang2017ADL}, S. Zhang showed that both large and small size replay buffers can significantly damage the performance. They proposed Combined Experience Replay (CER) to cope with this problem. It consists of adding the latest transition to the mini-batch pooled over the replay buffer which only requires $O(1)$ extra computation compares to PER to reduce the negative effect of the buffer size. The difference with PER is that by using CER, the latest transitions will undoubtedly be sampled. We decided to use the CER technique and to keep a mini-batch composed of randomly sample experiences $e_{rd\_n}$ and the last one as:
\begin{equation}
    \text{mini-batch}=[e_{rd\_1}\ ;\ e_{rd\_2}\ ;\ \dots\ ;\ e_{rd\_n-1}\ ;\ e_{-1}]
\end{equation}
with $n=256$, the size of our mini-batch. We kept a replay capacity of~$10^6$ as recommended in the original paper of the SAC~\cite{haarnojaSoft}.

\subsection{Application to the model-free controller} \label{mfcontroller}
We will now present the use of the previous deep reinforcement learning framework to build a model-free controller. The methodology follows the same line of thought as our previous work~\cite{TChaffre}. The objective of the learning agent is to build a predictive model, using the SAC algorithm, that directly maps the actions~\eqref{controlinputs} that control the MAV from the current state:
\begin{equation}
    \pi_\theta:\ s_t \rightarrow [\ \phi_\mathrm{r}\ ;\ \theta_\mathrm{r}\ ;\ \mathcal{T}\ ]
\end{equation}
where ($\phi_\mathrm{r}$,$\theta_\mathrm{r}$)$\in[-\frac{\pi}{6};+\frac{\pi}{6}]$ and $\mathcal{T}\in[m\times g;m\times (g+ 3.0)]$ (with $m=1.544$Kg, $g=9.81$m.s$^{-1}$). In order to achieve the target rallying mission, we need to provide relevant data to the agent. Therefore, we defined the following vector $o_t$ as the observation at the time-step $t$ of the environment:
\begin{equation} \label{eq:obs}
\begin{split}
o_t=[\ a_{t-1}\ ;\ \phi\ ;\ \theta\ ;\ \psi\ ;\ v_x\ ;\ v_y\ ;\ v_z\ ;\ \omega_\phi\ ;\ \omega_\theta\ ;\ \omega_\psi\ ;\ \zeta_t\ ;\ e_t\ ;\ d_t\ ]
\end{split}
\end{equation}

\noindent where $a_{t-1}$ are the last actions performed ; $[\phi;\theta;\psi]$ are the Euler angles of the robot (roll, pitch and yaw) ; $[v_x;v_y;v_z]$ and $[\omega_\phi;\omega_\theta;\omega_\psi]$ are its linear and angular velocities respectively in $\mathcal{R}_W$ and $\mathcal{R}_B$ ; $\zeta_t$ represents its position in $\mathcal{R}_W$ (obtained from its embedded odometry sensor), $e_t=[e_x;e_y;e_z]$ are the current errors on the set-points in terms of Euclidean distance and $d_t$ is the current Euclidean distance between the hexacopter and the target in $\mathcal{R}_W$. All the variables involved are assumed to be measured, their estimation is out of the scope of this work. We denote the target position by $\Lambda=[\Lambda_x;\Lambda_y;\Lambda_z]^T \in\mathcal{R}_W$. The dimension of this observation vector is $19$ and it has been standardized to have zero mean and a variance of $1$. In order to provide a higher time horizon to the agent, we constructed the state vector out of the current and past two observations vectors $[o_t;o_{t-1};o_{t-2}]$. For the purpose of providing the agent a sense of ``velocity" in the evolution of the state, we consider the two by two difference of these vectors such as $vel_t=(o_t-o_{t-1})$ and $vel_{t-1}=(o_{t-1}-o_{t-2})$. We went even further and tried to add a sense of ``acceleration" in the evolution of the state by including the difference between the latter vectors $acc_t=(vel_t-vel_{t-1})$. The resulting state vector is therefore defined as:
\begin{equation} \label{stateVEC1}
\begin{split}
s_t=[\ o_t\ ;\ o_{t-1}\ ;\ o_{t-2}\ ;\ vel_t\ ;\ vel_{t-1}\ ;\ acc_t\ ]
\end{split}
\end{equation}

The dimension of this state vector is $114$. The reward function $r_t$ has been designed in order to teach the agent how to complete the mission of target rallying:
\begin{equation} \label{reward_function}
r_t = \left\{
\begin{array}{l}
r_{receded} \ \hspace{5px} \textbf{ if } \ \ d_{rate} \leqslant 0  \\
r_{forward} \hspace{5px} \textbf{ if } \ \ d_{rate} > 0 \\
r_{reached} \ \ \hspace{1px} \textbf{ if } \ \  d_t \leqslant d_{reached} \\
r_{failed} \ \ \hspace{7px} \textbf{ if } \ \  z_w\notin[0.25;20]
\end{array}
\right.
\end{equation}

where $r_t$ is the reward of the agent at time $t$ ; $d_{rate}$ is the distance rate to the target performed between the last two steps such as ${d_{rate} = d_t - d_{t - 1}}$ ; $d_{reached}$ is the limit value beneath which we consider the target to be reached~; $z_w\in\mathcal{R}_W$ is the MAV altitude; both $r_{reached}$ and $r_{failed}$ are terminal rewards determined at the end of the ongoing episode. Each of these terms represent the specific features of the desired behavior of the MAV:
\begin{itemize}
    \item $r_{receded}$ is a constant negative reward equal to  $-20$ that is sent to the agent whenever the robot is getting away from the target (or staying immobile).
    \item $r_{forward}$ is a positive reward sent when the relative distance to the target is decreasing. It is defined as:
\begin{equation*}
    r_{forward}=C_1\times e^{\left(-\big[(\frac{d_t}{1+d_{rate}})\times\frac{1}{C_2}\big]^2\right)}
\end{equation*}

With this design, we encourage the robot to move toward the target as fast as possible. We chose the value of the constants $C_1$ in order to scale the positive reward signal. This is particularly important because as mentioned in~\cite{haarnojaSoft}, the SAC algorithm is particularly sensitive to the scaling of the reward signal which is the magnitude of the reward value. We followed the recommendation prescribed in~\cite{haarnojaSoft} and chose to set $C_1=20$ in order to obtain a positive reward scale of $20$ (which we found in practice to be the reward scale that gave us the best performance for this problem). The constant $C_2$ represents how sparse is $r_{forward}$ based on the distance to the target. We chose the value $C_2=20$ empirically. Therefore, the positive reward is equal to:
\begin{equation}
        r_{forward}=20\times e^{\left(-\big[(\frac{d_t}{1+d_{rate}})\times\frac{1}{20}\big]^2\right)}
\end{equation}

\item A constant positive reward is sent to the agent when it succeeded to complete the mission, meaning $d_t\leq d_{reached}$. This generates $r_{reached}=+1000$.
\item If the MAV altitude $z_w$ exceed a predefined threshold, the constant negative reward $r_{failed}=-550$ is generated.
\end{itemize}
We define as \textit{sampling rate}, the rate at which a state is sampled from the environment after the execution of the actions. A good practice consists in synchronizing it with the slowest sensor of the system which ensure no potential loss of information in the state vector and the fastest sampling rate. We synchronized it with the embedded odometry sensor which led to a \textit{sampling rate} of about 20Hz.

\subsection{Application to the learning-based controller} \label{lbcontroller}
As stated in Section \ref{learning_based}, learning-based methods consist in exploiting standard model-based control architectures that are either tuned or redesigned by a model-free algorithm in order to compensate for the unknown part of the model. 
The MAV is subject to an additive but unknown wind perturbation.
Therefore, we combined the PID controller~\eqref{controlLAW} under parametrization~\eqref{control_gains} within the DRL framework (Section \ref{SAC}) to iteratively auto-tune the feedback gains parameters (i.e. the constants $\tau_1$, $\tau_2$ and $\tau_3$), with the ambition of optimizing online a desired performance cost function. For behavior stability purpose, instead of directly trying to estimate the values of $\tau_i$, the SAC algorithm estimates at each time-step small increments $\nabla_i$ that are added to the $\tau_i$ in order to cope with the wind disturbances. By doing so, we avoid jumping from a configuration of gains to a totally different one, which usually give rise to undesired oscillations. The gains~\eqref{control_gains} are then computed using $\tau_i = \tau_i+\nabla_i$ with $\nabla_i\in[-0.01;+0.01]$ (while making sure that $3 \geq \tau_1 \geq \tau_2 \geq \tau_3 \geq 5\times10^{-3}$) and the resulting control law \eqref{controlLAW} is finally applied. The $\tau_i$ parameters are initialized to the nominal configuration $\tau_1=1,\tau_2=2.5,\tau_3=0.875$. The objective of the learning agent is now to build a predictive model that directly maps $\nabla_i$ from the current state to adjust the gains starting from this inital configuration:
\begin{equation}
\begin{split}
&\pi_\theta:\ s_t \rightarrow [\ \nabla_{\tau_{1_{roll}}}\ ;\ \nabla_{\tau_{2_{roll}}}\ ;\ \nabla_{\tau_{3_{roll}}}\ ;\ \nabla_{\tau_{1_{pitch}}}\ ;\\
&\nabla_{\tau_{2_{pitch}}}\ ;\ \nabla_{\tau_{3_{pitch}}}\ ;\ \nabla_{\tau_{1_{thrust}}}\ ;\ \nabla_{\tau_{2_{thrust}}}\ ;\ \nabla_{\tau_{3_{thrust}}}\ ]
\end{split}
\end{equation}

We used a slightly different observation vector from the model-free scheme \eqref{eq:obs}. Indeed, we propose to add the resulting parameters and the PID controller outputs \eqref{controlLAW} in the observation vector:
\begin{equation} \label{stateVEC2}
\begin{split}
o_t=&[\ a_{t-1}\ ;\ \tau_{rpt}\ ;\ pid_{rpt}\ ; \ k_{p_{roll}}\ ;\ k_{i_{roll}}\ ;\ k_{d_{roll}}\ k_{p_{pitch}}\ ;\ k_{i_{pitch}}\ ;\\&k_{d_{pitch}}\ ;\ k_{p_{thrust}}\ ;\ k_{i_{thrust}}\ ;\ k_{d_{thrust}}\ ;\ \phi\ ;\ \theta\ ;\ \psi\ ;\ v_x\ ;\ v_y\ ;\\&v_z\ ;\ \omega_\phi\ ;\ \omega_\theta\ ;\ \omega_\psi\ ;\ \zeta_t\ ;\ e_t\ ;\ d_t\ ]
\end{split}
\end{equation}
where $dim(\tau_{rpt})=9$ and $pid_{rpt}=[\phi_\mathrm{r} ;\ \theta_\mathrm{r}\ ;\ \mathcal{T}]$ . The dimension of this observation vector is $37$. We constructed the state vector out of this observation vector as earlier \eqref{stateVEC1}, resulting to a state vector of dimension $222$. We used the exact same reward function and parameters as described earlier \eqref{reward_function} to train this controller.

\section{Results} \label{results}
\begin{figure}[!h]
\captionsetup{justification=justified}
\begin{subfigure}{0.5\textwidth}
\centering
\includegraphics[scale=0.25]{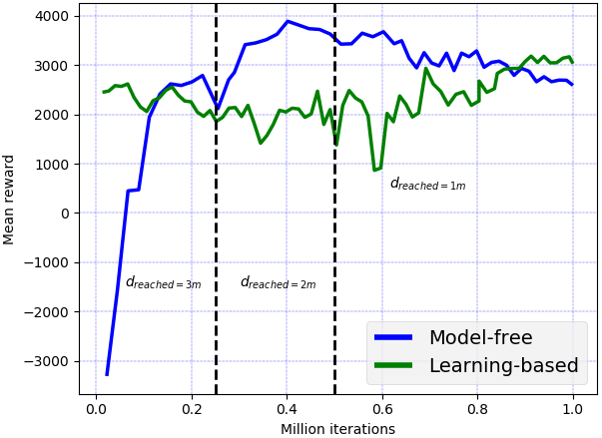} 
\caption{Mean reward}
\end{subfigure}
\begin{subfigure}{0.5\textwidth}
\centering
\includegraphics[scale=0.25]{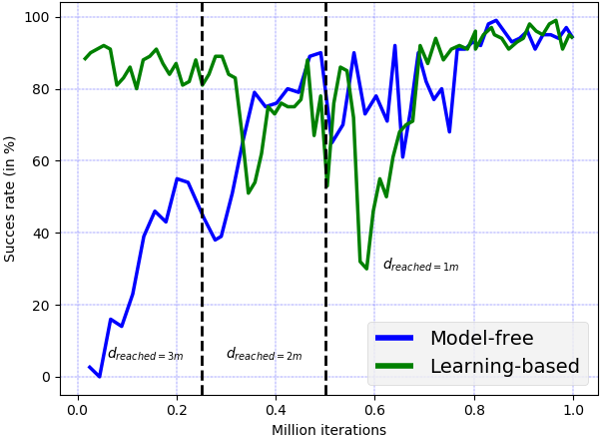}
\caption{Success rate}
\end{subfigure}
\caption{Training curves showing the mean reward and success rate computed per 100 episodes over a moving window of 100 episodes. Note that with SAC, the policy is trained to maximize also the entropy, therefore the mean action does not always correspond to the optimal action for the maximum return objective.}
\label{trainingcurves}
\end{figure}

A training episode is defined as follows: at the beginning of the episode, the MAV is set at the center of the environment at an altitude of $3$ meters with roll, pitch and yaw angles equal to $0$. A target is then initialized at a fixed and uniformly random position $\Lambda=[\Lambda_x;\Lambda_y;\Lambda_z]^T$ with $[\Lambda_x;\Lambda_y]^T\in[-20;-5[\cup]5;20]$ and $\Lambda_z\in[2;20]$. The mission then begins and is considered as a success if the relative distance to the target is inferior to a predefined threshold $d_{reached}$ and as a failure if an error signal is generated, both cases ending the episode. Otherwise, the episode is ended if the number of iteration steps reach the maximum value allowed per episode that is set at $300$. The training for both controllers consisted in performing $1\ 000\ 000$ iterations in a environment with a varying wind field (as described in Section~\ref{RotorS}). To help the agent, $d_{reached}$ is reduced during training as follows: at first $d_{reached}=3$m, from iteration $250\ 000$th we set $d_{reached}=2$m and from iteration $500\ 000$th we set $d_{reached}=1$m. The PyTorch framework~\cite{NEURIPS2019_9015} was used to carry out the numerical experiments, along with the CUDA toolkit~\cite{Nickolls2008ScalablePP} and an RTX 2060 GPU card, allowing us to perform the training of one controller in approximately $10$ hours.
\renewcommand{\thefootnote}{3}
It can be seen in Figure \ref{trainingcurves} that during training, both the learning-based (LB) and model-free (MF) controllers were able to reach a high success rate under unknown wind gust disturbances, which shows the applicability of the SAC DRL procedure for this type of aerial navigation problems. The LB strategy presents a much higher convergence speed to a significant success rate than the MF strategy, which shows the great potential of combining model-based classical controllers with learning procedures. This can be explained by the model-based part of the LB controller which allows it to choose relatively good actions despite being at the early stage of the training session. Therefore, from the beginning of the training the LB controller is able to explore a much higher part of the reward space than the one of the MF controller. We believe this significantly helps the DRL algorithm to find better overall strategies.
The evaluation consisted in performing the target rallying mission in areas of the same environment that had never been explored by neither controllers during training (i.e. the wind field in these areas were totally unknown to the neural networks) with $d_{reached}=1$m. The evaluation is composed of a total of $500$ episodes, different from each other in terms of target position and with a max step size per episode of $1000$. The targets during evaluation were uniformly distributed in the space defined by $\Lambda=[\Lambda_x;\Lambda_y;\Lambda_z]^T$ with $[\Lambda_x;\Lambda_y]^T$ $\in[-50;-20[\cup]20;50]$ and $\Lambda_z$ $\in[2;20]$. The same set of evaluation episodes was used for each controller. We also evaluated a fixed control strategy which consisted in a PID controller with the fixed nominal poles configuration (\ref{lbcontroller}). The outcomes of this evaluation are provided in Table \ref{results1}. We can see that the LB controller achieves a better performance for each metric. Noted that it requires, on average, less actions to achieve the task with the LB controller despite sharing the same \textit{sampling rate}. The mean reward per step of the LB controller is more than $2$ times higher than the MF one. On the other hand, the fixed control strategy is showing a much lower success rate, close to $50\%$. We observed that with this strategy, failures mostly consisted in cases where the MAV is close to the steady-state but is being deviated by the wind gust. The vehicle is then not able to recover with fixed-pole parametrization. It shows the benefits of adaptive control methods when facing unknown disturbances.\vspace*{-.5cm}
\vspace*{-1.45cm}
\begin{table}[!h]
\centering
{\renewcommand{\arraystretch}{1}
\begin{tabular}{|P{2.5cm}||P{1.75cm}|P{1.75cm}|P{2cm}|P{1.5cm}|P{1.75cm}|}
  \hline
  \thead{Controller type} & \thead{Mean step\\number} & \thead{Mean total\\reward} & \thead{Mean reward\\per step} & \thead{Success\\rate} &  \thead{Positive\\reward rate}\\
  \hline \hline
  Fixed Poles PID & 488 & 710.797 & 1.454 & 50.6\% & 61.197\%\\
  \hline
  Model-free & 357 & 2238.970 & 6.256 & 74.2\% & 86.056\%\\
  \hline
  Learning-based & \textbf{281} & \textbf{4034.434} & \textbf{14.346} & \textbf{91.6\%} & \textbf{89.2\%} \\
  \hline
\end{tabular}}
\vspace{0.25cm}
\caption{Evaluation results.\label{results1}}
\end{table}

\section{Conclusions} \label{discussion}
We present a novel application of the SAC procedure for learning both model-free and learning-based adaptive controllers applied to the autonomous navigation of a MAV under wind gust conditions. These two strategies have been trained and compared in the same ROS-Gazebo reference simulation. Both strategies were able to reach high levels of performances under unknown uncertainty conditions, but the learning-based scheme, with a judicious parametrization of its model-based control part, exhibited a much faster convergence rate. Our results suggest that learning-based adaptive methods can be much more efficient than model-free ones and allow a better stability analysis in the DRL scheme.
\renewcommand{\abstractname}{Acknowledgements}\vspace*{-0.25cm}
\begin{abstract}
The authors thank Dr. Estelle Chauveau from Naval Group for the help provided. This work was supported by SENI, the research laboratory between Naval Group and ENSTA Bretagne.
\end{abstract}

\end{document}